\documentclass[preprint,10pt]{elsarticle}

\usepackage{amssymb}
\usepackage{amsmath}

\usepackage{graphicx}
\usepackage{subcaption}
\usepackage{soul}
\usepackage{color}
\usepackage{amsmath}
\usepackage{bbm}
\usepackage{multirow}
\usepackage{makecell}
\usepackage{pifont}
\usepackage{booktabs}
\usepackage{hyperref}

\usepackage{algorithmic}
\usepackage{algorithm}
\usepackage{amssymb}
\usepackage{multirow}
\usepackage{makecell}
\usepackage{arydshln}
\usepackage{pifont}
\usepackage{lscape}
\usepackage{soul}

\usepackage[dvipsnames, svgnames, x11names]{xcolor}

\biboptions{sort&compress}

\makeatletter

\newcommand{\Rmnum}[1]{\expandafter\@slowromancap\romannumeral #1@}
\makeatother

\newtheorem{definition}{Definition} 

\journal{Pattern Recognition}

\begin{document}

\begin{frontmatter}



\title{MCE: Towards a General Framework for Handling Missing Modalities under Imbalanced Missing Rates\footnote{This is the accepted version of an article that has been published in \textbf{Pattern Recognition}. The final version is available via the DOI, or for 50 days' free access via this Share Link: \url{https://authors.elsevier.com/a/1m40D77nKsBm-} (valid until December 28, 2025).}
} 




\author{Binyu Zhao, Wei Zhang, Zhaonian Zou \\
\textit{The School of Computer Science and Technology, Harbin Institute of Technology, Harbin, 150001, China}
}

\begin{abstract}
Multi-modal learning has made significant advances across diverse pattern recognition applications. However, handling missing modalities, especially under imbalanced missing rates, remains a major challenge. This imbalance triggers a vicious cycle: modalities with higher missing rates receive fewer updates, leading to inconsistent learning progress and representational degradation that further diminishes their contribution. Existing methods typically focus on global dataset-level balancing, often overlooking critical sample-level variations in modality utility and the underlying issue of degraded feature quality. We propose Modality Capability Enhancement (MCE) to tackle these limitations. MCE includes two synergistic components: i) Learning Capability Enhancement (LCE), which introduces multi-level factors to dynamically balance modality-specific learning progress, and ii) Representation Capability Enhancement (RCE), which improves feature semantics and robustness through subset prediction and cross-modal completion tasks. Comprehensive evaluations on four multi-modal benchmarks show that MCE consistently outperforms state-of-the-art methods under various missing configurations. The final published version is now available at \url{https://doi.org/10.1016/j.patcog.2025.112591}. Our code is available at \url{https://github.com/byzhaoAI/MCE}.
\end{abstract}



\begin{keyword}
Incomplete multi-modal learning \sep
imbalanced missing rate \sep
capability enhancement


\end{keyword}

\end{frontmatter}



\section{Introduction}
\label{sec:intro}
Multi-modal learning has recently driven progress in pattern recognition (PR) across areas such as medical diagnosis~\cite{he2023co}, computational social science~\cite{wang2023tetfn,liu2024sarcasm}, and autonomous systems~\cite{cao2024madtp,li2025multimodal}. Generally, by fusing complementary information from different modalities, multi-modal methods outperform single-modality approaches~\cite{baltruvsaitis2018multimodal,bayoudh2022survey}. Nevertheless, due to sensor failures, acquisition costs, or privacy restrictions, one or more modalities may be missing in real-world applications, which frequently complicates model training and substantially hinders model deployment.

Existing works have explored incomplete multi-modal learning~\cite{zhao2022modality,chen2024novel,yang2023learning,wang2023multi,lian2023gcnet,xiong2025disentanglement}. However, an important and underexplored problem still remains: imbalanced missing rates across modalities. Consider a medical diagnosis task where MRI scans (modality A) might be unavailable 60\% of the time, whereas CT scans (modality B) are missing only 10\%. A model trained on such data tends to over-rely on the more frequently available CT scans. As a result, it performs poorly when MRI data is absent. and fails to fully exploit complementary information when both modalities are available, since it has not sufficiently learned MRI-specific features.

This imbalance produces a reinforcing cycle of \textbf{model bias} and \textbf{representational degradation}. During training, modalities with higher missing rates contribute fewer gradient updates, so their parameters receive less learning signal and converge more slowly. This imbalance gradually biases parameter updates and weakens the model’s capacity to extract informative features from under-represented modalities. The result is poorer representations for those modalities and further reduction in their utility, which exacerbates the original bias. Therefore, solving this problem requires interventions that target both the learning dynamics and the quality of representations.

Some recent methods attempt to correct imbalance using dataset-level discrepancy measures~\cite{sun2024redcore, shi2024passion}. However, they typically assume uniform modality gaps across samples and do not explicitly improve representation quality. Besides, they are also designed for particular domains and do not generalize well. In practice, both modality discrepancies and representation quality vary at the sample level~\cite{wei2024enhancing}, motivating a more general framework that can i) balance learning progress across modalities and ii) enhance the representational power of under-represented modalities, regardless of their missing rates.

We introduce the Modality Capability Enhancement (MCE) framework to tackle these challenges. MCE combines two complementary components, Learning Capability Enhancement (LCE) and Representational Capability Enhancement (RCE). LCE alleviates procedural imbalance by adaptively adjusting gradient incentives: a dataset-level factor captures global modality availability, while a batch-level factor—derived from Shapley values—assesses each modality’s current learning state and allocates differentiated incentives. RCE counteracts representational degradation by improving feature quality. RCE comprises a subset-combination prediction task that aligns predictions from modality subsets with the ground truth via ensemble-style learning, together with a self-supervised completion objective that reconstructs missing modalities omitted by the subsets to encourage richer cross-modal interactions. These objectives promote robust, task-relevant feature learning across all modality subsets.

Crucially, LCE and RCE work in concert: LCE ensures fair learning opportunities for each modality while RCE ensures those opportunities yield high-quality representations. By jointly optimizing learning dynamics and representation quality, MCE breaks the vicious cycle caused by imbalanced missing rates and delivers a principled, practical solution. Comprehensive experiments on diverse PR tasks show that MCE consistently outperforms state-of-the-art (SOTA) baselines, highlighting robustness and broad applicability.

Our main contributions are listed as follows:
\begin{itemize}
\item \textbf{Paradigm shift}: We identify imbalanced missing rates as a general challenge in pattern recognition, not merely a domain-specific issue. And we trace its root causes to imbalanced learning dynamics and representational degradation.
\item \textbf{Principled quantification}: We introduce a Shapley value–based measure of modality contribution with a novel utility function and a practical performance bound for imbalance analysis.
\item \textbf{Synergistic integration}: We propose MCE, which unifies dynamic learning incentives (LCE) with representation enhancement (RCE), and demonstrate that their integration is critical for robust multi-modal learning.
\item \textbf{Empirical effectiveness}: We validate MCE across multiple PR tasks and establish a strong, generalizable baseline for future research.
\end{itemize}


\section{Related Works}
\label{sec:related}

\subsection{Imbalanced Multi-modal Learning}
\label{sec:related1}
Multi-modal learning inherently encounters fairness and imbalance challenges arising from modality heterogeneity~\cite{zhang2021assessing, fernando2021missing}. Models tend to bias toward stronger or more informative modalities, which underutilizes others and degrades overall performance~\cite{peng2022balanced, huang2022modality}. To mitigate this, prior work has proposed techniques such as prototype alignment to strengthen weaker modalities~\cite{fan2023pmr}, adaptive feature- and modality-weighting strategies~\cite{nguyen2024ada2i}, and dynamic allocation of training weights according to modality differences~\cite{wu2025balanced}. These approaches are effective at addressing intrinsic modality gaps but typically assume complete data. Therefore they do not handle the additional complications introduced by missing modalities or sample-level noise.

Wei \textit{et al.} introduced a sample-level assessment for missing data, defining a value function $v(\cdot)$ based on available modalities to guide resampling~\cite{wei2024enhancing}. While useful for diagnosing imbalance, this resampling strategy can induce bias. Their follow-up gradient-based approach—tracking learning progress via clustering purity—entails substantial computational cost~\cite{wei2024diagnosing}. Crucially, these methods focus on balancing training dynamics or diagnosing imbalance, rather than directly improving degraded feature representations, which is central to our work.

\subsection{Incomplete Multi-modal Learning}
\label{sec:related2}
Incomplete multi-modal learning targets scenarios where some modalities are missing because of hardware limits, cost, or privacy. Existing methods fall roughly into two categories. \textit{Reconstruction-based} methods impute missing data through generative models or cross-modal transfer prior to fusion~\cite{zhao2022modality, chen2024novel, yang2023learning, huan2023unimf, lian2023gcnet}. Examples include Transformer-based patch reconstruction~\cite{chen2024novel} and encoder–decoder completion networks~\cite{yang2023learning}. \textit{Analysis-based} methods avoid explicit reconstruction and instead focus on filtering irrelevant components or extracting shared features for fusion~\cite{reza2024robust, wang2024unibev, wang2023multi, xiong2025disentanglement}. Representative techniques include substituting missing-modality inputs with shared features~\cite{wang2023multi} or disentangling common and modality-specific features to improve fusion~\cite{xiong2025disentanglement}.

A common limitation across these works is the assumption of random missingness. They are not designed for the more challenging case of \textit{imbalanced} missing rates, where systematic absence of particular modalities induces persistent bias and harms representation learning.

\subsection{The Intersection: Imbalanced Missing Rates}
\label{sec:related3}
When intrinsic modality heterogeneity and extrinsic missingness coexist, the result is a compounded imbalance. This is harder to address than either issue in isolation. Methods that target inherent modality gaps typically fail under systematic missingness, while standard incomplete-learning techniques that assume random patterns struggle when missingness is imbalanced.

Recent efforts have begun to design heuristic weighting schemes to tackle this intersection. Sun \textit{et al.} proposed a bi-level optimization that uses a “relative advantage” metric to adapt supervision dynamically~\cite{sun2024redcore}. Shi \textit{et al.} combined self-distillation with gradient regularization to balance convergence rates according to modality preferences~\cite{shi2024passion}. These contributions focus on dataset-level adjustments to global convergence, but they neither explicitly address sample-level variation in modality utility, nor mitigate the degradation of features for frequently missing modalities. Closing this gap—balancing learning dynamics while improving representation quality at the sample level—is the core aim of our MCE framework.

\subsection{Shapley Value}
\label{sec:related4}
The Shapley value, from cooperative game theory, provides a principled way to allocate payoffs among players according to their marginal contributions across all coalitions~\cite{chalkiadakis2011computational, fatima2008linear}. Due to its axiomatic properties (efficiency, symmetry, dummy, additivity), it has been adopted in machine learning for feature selection and model interpretation.

\begin{definition}[Shapley Value]
A cooperative game is defined by a set of players $\mathcal{M}={1,2,...,M}$ and a characteristic function $v: 2^M \rightarrow \mathcal{R}$, which assigns a value to each subset of the coalition $\mathcal{S} \subseteq \mathcal{M}$, representing the total payoff that the subset $\mathcal{S}$ can achieve. The Shapley value of the $i$-th player is given by:
\begin{equation}
    \phi_i = \sum_{\mathcal{S} \subseteq \mathcal{M} \setminus \{i\}} \frac{|\mathcal{S}|!(|\mathcal{M}|-|\mathcal{S}|-1)!}{|\mathcal{M}|!}\Phi_i(\mathcal{S})
\end{equation}
where $\Phi_i(\mathcal{S}) = v(\mathcal{S} \cup \{i\}) - v(\mathcal{S})$ is the marginal contribution, which measures how
 much value the $i$-th player attributes to the subset $\mathcal{S}$.
\end{definition}

Despite its appeal, Shapley-based techniques in multi-modal settings have mainly been used for post-hoc explanation or with simple value functions (\textit{e.g.}, model performance for a subset). These applications do not account for changing learning states or the degradation of representations under imbalanced missingness. In contrast, we employ the Shapley value within training as a dynamic incentive mechanism. We also design a new utility function and a practical performance bound. They make Shapley computations tractable and useful for adjusting modality incentives during learning.


\begin{figure}[!t]
\centering
\includegraphics[width=\linewidth]{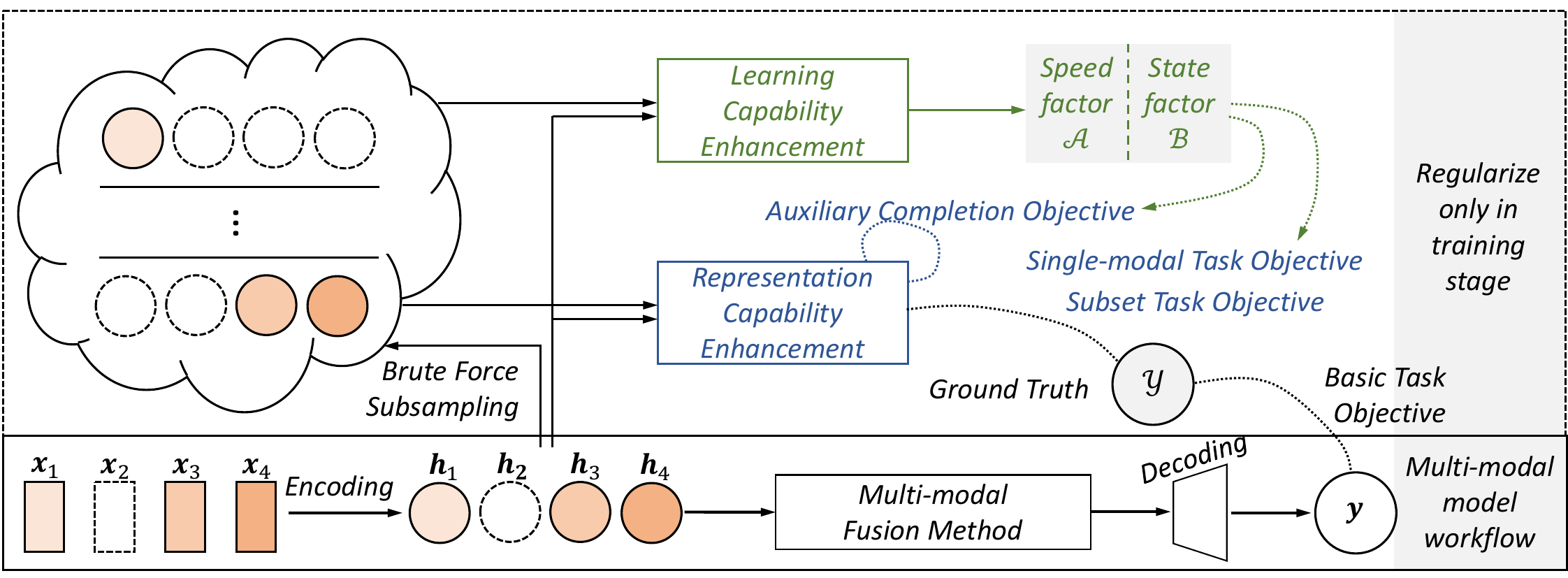}
\caption{Workflow of a multi-modal model and overview of the proposed MCE framework. During training on data with imbalanced missing rates, MCE incorporates Learning Capability Enhancement (LCE, in {\color{YellowGreen}{green}}) to balance modality-specific learning progress and encourage each modality to reach its performance potential, and Representation Capability Enhancement (RCE, in {\color{RoyalBlue}{blue}}) to enrich representation semantics by exposing the model to a wider variety of multi-modal combinations.}
\label{fig:arch}
\end{figure}

\section{Method}
\label{sec:method}

\subsection{Problem Formulation}
We consider an incomplete multi-modal dataset $\mathcal{D}=\{(\mathcal{X}_n,\mathcal{E}_n,\mathcal{Y}_n)\}_{n=1}^N$. For each sample $n$, $\mathcal{X}_n=\{\mathbf{x}_{n,m}\}_{m=1}^M$ contains up to $M$ modalities, and $\mathcal{E}_n=\{\mathcal{E}_{n,m}\}_{m=1}^M$ is a binary presence indicator with $\mathcal{E}_{n,m} \in \{0,1\}$. $\mathcal{Y}_n$ denotes the task-specific gournd truth.

Samples for which all modalities are absent ($\sum_m \mathcal{E}_{n,m} = 0$) are excluded from training and evaluation, because they carry no observable signal. This standard practice avoids introducing unlearnable noise. Handling such cases at deployment (\textit{e.g.}, via population priors or external predictors) is beyond the scope of this work.

Our model follows the standard encode-fuse-decode pipeline. For each available modality $m$, an encoder $f_{enc}^m(\cdot)$ produces a hidden feature $\mathbf{h}_{n,m} \in \mathcal{R}^{C \times D}$, where $C$ is the channel dimension and $D$ the spatial/temporal dimension (1D/2D/3D as appropriate). A fusion module $f_{fusion}(\cdot)$ aggregates modality features into a joint representation $\mathbf{h}_n \in \mathcal{R}^{C \times D}$, which a decoder $f_{dec}(\cdot)$ maps to the prediction $\mathbf{y}_n$:
\begin{equation}
\begin{split}
    \mathbf{h}_{n,m} &= f_{enc}^m(\mathbf{x}_{n,m}), \quad m=1,2,...,M \\
    \mathbf{h}_n &= f_{fusion}(\{\mathbf{h}_{n,m}\}_{m=1}^M) \\
    \mathbf{y}_n &= f_{dec}(\mathbf{h}_n) \\
\end{split}
\end{equation}

The primary training objective minimizes a task-specific loss $\mathcal{L}_{task} = \varrho(\mathbf{y}_n, \mathcal{Y}_n)$ (\textit{e.g.}, cross-entropy or mean squared error). MCE augments this pipeline to improve robustness under imbalanced missing rates. It operates at the encoder, incentive, and representation levels while remaining \textit{agnostic} to the choice of $f_{fusion}(\cdot)$. An overview is provided in Fig.~\ref{fig:arch}.

\subsection{Learning Capability Enhancement (LCE)}
\label{method:lce}
Imbalanced missing rates cause both unequal data exposure and skewed learning dynamics. Whereas prior methods introduced dataset-level corrections to balance supervision~\cite{sun2024redcore, shi2024passion}, they frequently ignore sample-level variation in modality utility. LCE addresses imbalance at two scales: a dataset-level correction compensates for systematic differences in modality availability, and a batch-level, Shapley value–based incentive adapts training signals to the sample-specific learning state.

\begin{figure}[!t]
\centering
\includegraphics[width=\textwidth]{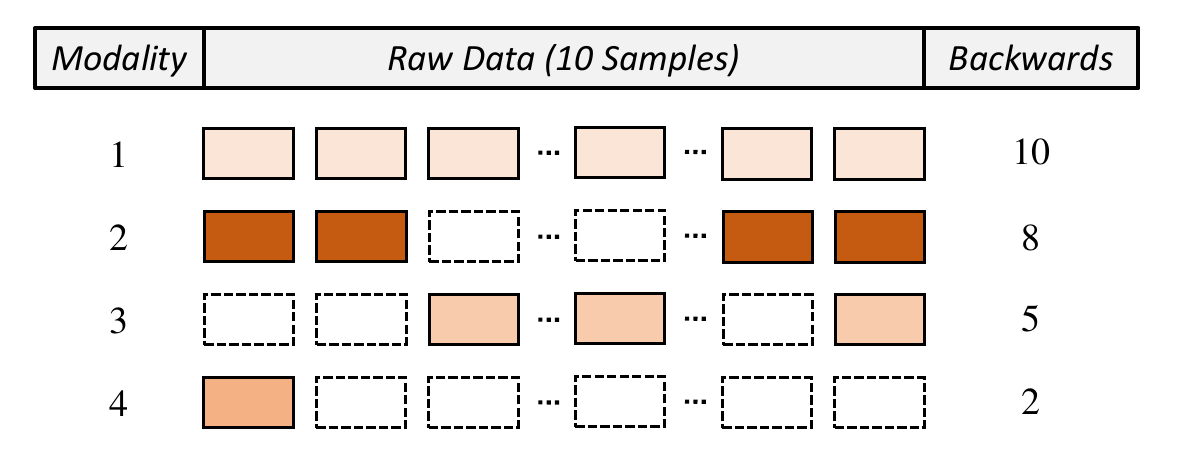} 
\caption{A 4-modality, 10-sample dataset example to explain the rationale for introducing dataset-level modal rating. When the ratings of modalities are identical, update (backward) times of different modality networks in the multi-modal model may exhibit significant disparities (10, 8, 5, 2 in the example).}
\label{fig:rating}
\end{figure}

\subsubsection{Missing rate balancing}
Imbalanced missing rates lead to unequal update frequencies and bias the optimizer toward dominant modalities (see Fig.~\ref{fig:rating}). To correct this procedural bias, we define a global speed factor $\mathcal{A} \in \mathcal{R}^M$:
\begin{equation}
\label{eq:factorA}
    \mathcal{A}_m=\frac{N}{\sum_{n=1}^N \mathcal{E}_{n,m}}
\end{equation}

When modality $m$ is present, its modality-specific loss (or gradient) is scaled by $\mathcal{A}_m$, so rarer modalities produce proportionally larger training signals. This reweighting enforces update-level fairness and is computationally simple compared with nested bi-level schemes (e.g.,\cite{sun2024redcore}).

To avoid instability from extreme scaling, we normalize $\mathcal{A}$ (for example, by its mean) before applying it. And we combine this dataset-level correction with the batch-level, Shapley-derived incentives described next for sample-aware adaptation.

\begin{figure}[!t]
\centering
\includegraphics[scale=0.46]{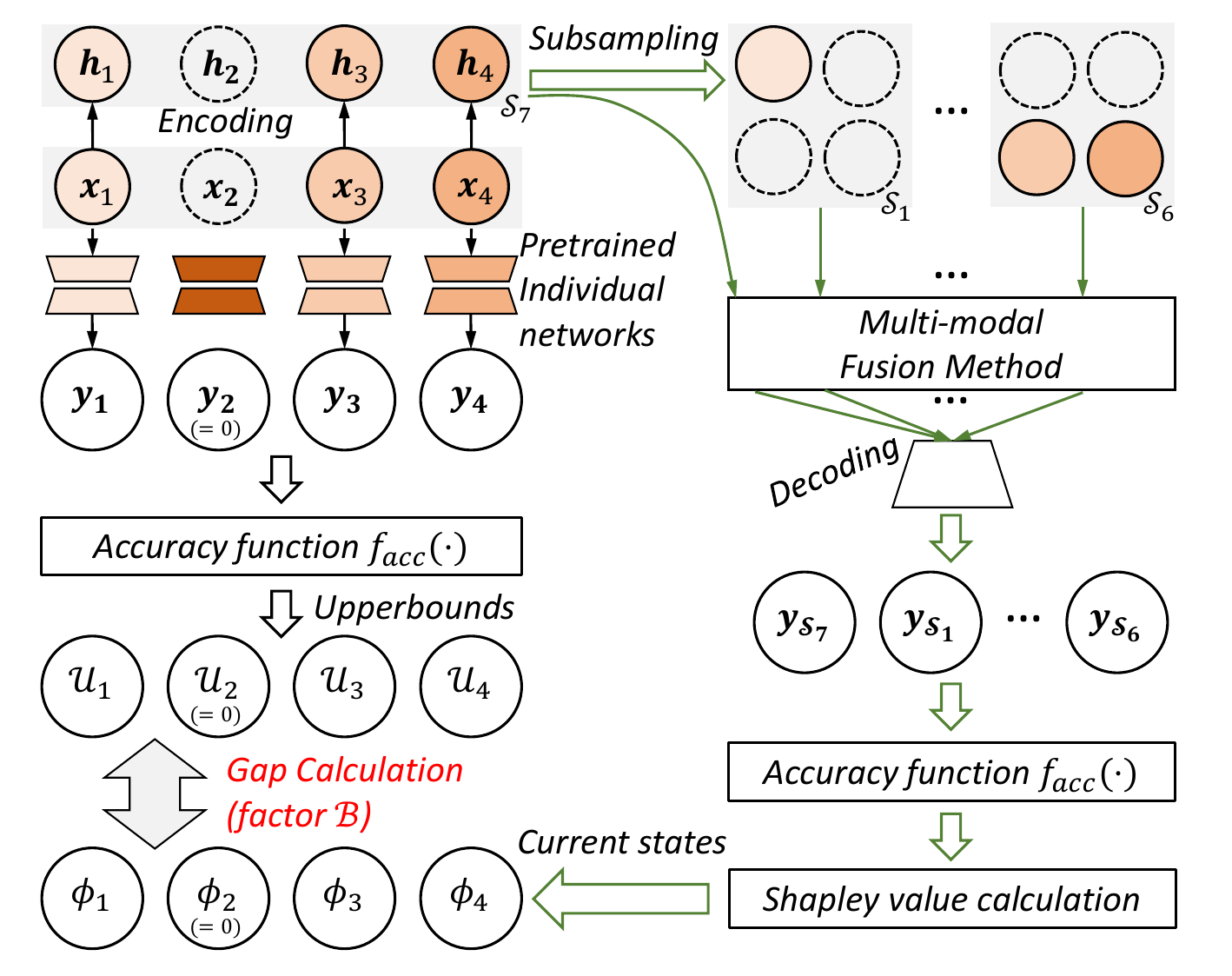} 
\caption{A example of batch size 1 with 4-modality sample (missing the 2nd modality) to generate factor $\mathcal{B}$ for learning state encouraging.}
\label{fig:shap}
\end{figure}

\subsubsection{Learning state encouragement}
Equalizing update counts does not guarantee a modality reaches its performance potential. To measure unrealized potential in a sample-aware way, we introduce a factor $\mathcal{B} \in \mathcal{R}^{M}$ based on Shapley value marginal contributions.

For a batch of $N_b$ samples and the available-modality set $\mathcal{M}_n$ for sample $n$, define the coalition value of subset $\mathcal{S} \subseteq \mathcal{M}_n$ as the accuracy obtained using only modalities in $\mathcal{S}$:
\begin{equation}
v_n(\mathcal{S}) = f_{acc}(f_{dec}(f_{fusion}(\{\mathbf{h}_{n,m}\}_{m \in \mathcal{S}})),\mathcal{Y})
\end{equation}

Here $f_{acc}(\cdot,\cdot)$ returns a per-sample accuracy appropriate to the task (\textit{e.g.}, 0/1 for classification, IoU/Dice for segmentation, or a tolerance-based indicator for regression). The batch Shapley value of modality $m$ is its average marginal contribution across all subsets:
\begin{equation}
\label{eq:shap}
\phi_m = \sum_{n=1}^{N_b} \sum_{\mathcal{S} \subseteq \mathcal{M}_n\setminus{\{m\}}} \frac{|\mathcal{S}|!(|\mathcal{M}_n|-|\mathcal{S}|-1)!}{|\mathcal{M}_n|!}[v_n(\mathcal{S}\cup{m})-v_n(\mathcal{S})]
\end{equation}

The Shapley value $\phi_m$ captures current contribution but not the modality’s achievable performance. We therefore define a per-modality upperbound $\mathcal{U}_m$ as the accuracy of a standalone, pre-trained single-modal model (kept frozen during multi-modal training):
\begin{equation}
\label{eq:upperbound}
\mathcal{U}_m = f_{acc}(y_m^{pt},\mathcal{Y}), \qquad
y_m^{pt} = f_{dec}^{m,pt}(f_{enc}^{m,pt}(\mathbf{x}_m))
\end{equation}

The encoder $f_{enc}^{m,pt}$ and decoder $f_{dec}^{m,pt}$ are trained with corresponding modality data in dataset $\mathcal{D}$. Using frozen single-modal models yields a stable, unbiased estimate of modality-specific potential and avoids confounds introduced by joint multi-modal training. Therefore, the upperbound $\mathcal{U}$ is a stable, empirical benchmark rather than a strict theoretical maximum. It represents the best performance one could realistically expect from a modality when learned in isolation on the available dataset. It provides a fixed target against which to measure the under-utilization of that modality within the multi-modal model.

\begin{definition}[Capability gap]
The capability gap for modality $m$ is the difference between its upperbound and current contribution:
\begin{equation}
    \Delta_m = \mathcal{U}_m - \phi_m
\end{equation}
\end{definition}

We convert this gap into a batch-normalized learning-state factor:
\begin{equation}
\label{eq:factorB}
    \mathcal{B}_m = \frac{1_{\Delta_m > 0}}{\sum_{n=1}^{N_b} \mathcal{E}_{n,m}} \Delta_m = \frac{1_{\Delta_m > 0}}{\sum_{n=1}^{N_b} \mathcal{E}_{n,m}} (\mathcal{U}_m - \phi_m)
\end{equation}

A large $\mathcal{B}_m$ indicates that modality $m$ is both under-observed and under-performing relative to its potential, and therefore merits stronger training incentives. Together, $\mathcal{A}$ and $\mathcal{B}$ form a two-tier mechanism: $\mathcal{A}$ corrects global update imbalance while $\mathcal{B}$ supplies sample-aware encouragement (illustrated in Fig.~\ref{fig:shap}).

It it worth noting that any negative $\mathcal{B}_m$ values are masked to zero in practice, meaning the model encourages learning but does not penalize modalities.

\subsection{Representation Capability Enhancement (RCE)}
\label{method:rce}
LCE identifies under-trained modalities and balances update frequency, but it does not by itself improve feature quality. Focus solely on training balance can still leave representations for under-represented modalities weak. RCE directly targets to produce robust, interoperable representations across modality combinations.

\textbf{Single-modal supervision.}
Each modality encoder receives direct, modality-specific supervision that is weighted by the LCE-derived signals to prioritize modalities that are both rare and under-performing:
\begin{equation}
\mathcal{L}_{single} = \sum_{m=1}^M \mathcal{A}_m \mathcal{B}_m \varrho(f_{dec}^m(\mathbf{h}_m), \mathcal{Y})
\end{equation}

The factors $\mathcal{A}_m$ and $\mathcal{B}_m$ adaptively amplify supervision for modalities, as they identify which modalities need to be focused more for stronger training.

\textbf{Subset-task supervision.}
To make representations robust across arbitrary modality combinations, we train the model to predict correctly from any non-empty subset of available modalities in the batch. For a batch of $N_b$ samples, let $\mathcal{M}_n^{exist}$ denote the available modalities for sample $n$. The set of non-empty subsets present in the batch is
\begin{equation}
\mathcal{S}_{batch} = \{\mathcal{S} \subseteq \mathcal{M} \setminus \{\varnothing\} \ | \ \exists \ n: \mathcal{S} \subseteq \mathcal{M}_n^{exist} \}
\end{equation}

For each $\mathcal{S} \in \mathcal{S}_{batch}$, let $\mathcal{N}_{\mathcal{S}}=\{n:\mathcal{S} \subseteq \mathcal{M}_n^{exist} \}$ be sample indices where subset $\mathcal{S}$ is feasible. The subset supervision loss is
\begin{equation}
    \mathcal{L}_{sub} = \frac{1}{|\mathcal{S}_{batch}|}\sum_{\mathcal{S} \in \mathcal{S}_{batch}} \frac{1}{|\mathcal{N}_{\mathcal{S}}| + \epsilon} \sum_{n\in\mathcal{N}_{\mathcal{S}}}\varrho(f_{dec}(f_{fusion}(\{\mathbf{h}_{n,m}\}_{m \in \mathcal{S}})), \mathcal{Y})
\end{equation}
where $\epsilon>0$ prevents division by zero. Averaging across all occurring subsets encourages each encoder to learn features that are effective both individually and in combination. It reduces the feature redundancy and increases information complementarity, finally  improving robustness to missing modalities.

\textbf{Auxiliary completion supervision.}
We also require the model to reconstruct features for all dropped modalities for every non-empty subset $\mathcal{S}$ occurring in the batch. This enforces that the latent space supports cross-modal inference from any available combination. For sample $n\in\mathcal{N}_{\mathcal{S}}$, the dropped modalities are $\mathcal{S}_{drop}^{(n,\mathcal{S})} = \mathcal{M}^{exist} \setminus \mathcal{S}$. Given the available features $\{\mathbf{h}_{n,k}\}_{k\in\mathcal{S}}$, the reconstruction module produces
\begin{equation}
    \mathbf{h}_{n,m}^{'} = f_{rec}(\{\mathbf{h}_{n,k}\}_{k \in \mathcal{S}},\mathcal{S},n) \quad m \in \mathcal{S}_{drop}^{(n,\mathcal{S})}
\end{equation}

The exhaustive auxiliary loss averages reconstruction error over all subsets and feasible samples, weighting each modality by the LCE factors:
\begin{equation}
    \mathcal{L}_{aux} = \frac{1}{|\mathcal{S}_{batch}|} \sum_{\mathcal{S} \in \mathcal{S}_{batch}} \frac{1}{|\mathcal{N}_{\mathcal{S}}|} \sum_{n\in\mathcal{N}_{\mathcal{S}}} \frac{1}{|\mathcal{S}_{drop}^{(n,\mathcal{S})}| + \epsilon} \sum_{m\in\mathcal{S}_{drop}^{(n,\mathcal{S})}} \mathcal{A}_m \mathcal{B}_m||\mathbf{h}_m^{'} - \mathbf{h}_m||
\end{equation}
Weighting the reconstruction loss by $\mathcal{A}_m$ and $\mathcal{B}_m$ focuses recovery on modalities that most need representation improvement.

Integrating all signals yields the final training objective:
\begin{equation}
\label{eq:total}
\mathcal{L} = \mathcal{L}_{task} + \lambda_{single}\mathcal{L}_{single} + \lambda_{sub}\mathcal{L}_{sub} + \lambda_{aux}\mathcal{L}_{aux}
\end{equation}

\begin{algorithm}[t]
\renewcommand{\algorithmicrequire}{\textbf{Input:}}
\renewcommand{\algorithmicensure}{\textbf{Output:}}
\caption{Modality Capability Enhancement Algorithm}
\label{alg:mce}
\begin{algorithmic}[1]
\REQUIRE Incomplete multi-modal dataset $\mathcal{D} = {(\mathbf{X}n, \mathcal{E}n, \mathcal{Y}n)}_{n=1}^N$
\ENSURE Optimized multi-modal model
\STATE Initialize model params $\theta$, independent pre-trained unimodal models (frozen)
\STATE Compute global factor $\mathcal{A}$ via Eq.~\ref{eq:factorA}
\FOR{epoch = 1 to T}
\FOR{batch in $\mathcal{D}$}
\STATE Get unimodal upperbounds $\mathcal{U}$ from frozen models (Eq.~\ref{eq:upperbound})
\STATE Compute Shapley values $\phi$ and factor $\mathcal{B}$ (Eqs.~\ref{eq:shap}, \ref{eq:factorB})
\STATE Compute total loss $\mathcal{L}$ (Eq.~\ref{eq:total})
\STATE Update $\theta \gets \theta - \eta \nabla_\theta \mathcal{L}$ \hfill $\triangleright$ \textbf{Key: Joint optimization}
\ENDFOR
\ENDFOR
\RETURN Optimized multi-modal model.
\end{algorithmic}
\end{algorithm}

\subsection{Synergistic Interaction: Breaking the Vicious Cycle}
\label{sec:synergy}
The principal novelty of MCE is the deliberate, closed-loop interaction between its two modules, which together break the vicious cycle of model bias and representational degradation described in Sec.~\ref{sec:intro}. This interaction is an iterative process executed within each training batch and can be understood as a diagnosis–treatment loop:

\textbf{Diagnose and incentivize.} LCE first diagnoses imbalance at two scales. The dataset-level factor $\mathcal{A}$ compensates for systematic differences in modality availability by increasing the learning signal for rarer modalities when they appear. The batch-level Shapley-derived factor $\mathcal{B}$ provides a fine-grained, sample-aware diagnosis. It measures each modality’s current contribution $\phi_m$ relative to its unimodal upperbound $\mathcal{U}_m$, yielding the capability gap $\Delta_m$. Together, $\mathcal{A}_m$ and $\mathcal{B}_m$ quantify where learning resources should be focused.

\textbf{Treat and enhance.} RCE uses LCE’s diagnosis as actionable guidance. The factors $\mathcal{A}$ and $\mathcal{B}$ weight representation objectives (\textit{i.e.}, $\mathcal{L}_{single}$ and $\mathcal{L}_{aux}$), so that modalities that are both rare and under-performing receive proportionally stronger supervision. The subset prediction loss $\mathcal{L}_{sub}$ complements this by enforcing that enhanced features remain robust and interoperable across arbitrary modality combinations.

These components form a positive feedback loop during training. By increasing effective update signals, LCE creates the conditions for under-observed modality encoders to learn. RCE then converts those extra updates into high-quality, task-relevant features through targeted supervision and cross-modal reconstruction. As representation quality improves, a modality’s Shapley value $\phi_m$ increases, its capability gap $\Delta_m$ shrinks, and its batch incentive $\mathcal{B}_m$ is reduced in subsequent iterations. Over time this iterative process narrows capability gaps and balances both learning dynamics and representational power across modalities.

In short, LCE opens the learning channel for under-represented modalities and RCE fills that channel with high-quality supervision. Their joint operation is more effective than either alone; empirical ablations in Sec.~\ref{sec:ablation} demonstrate that combining LCE and RCE yields substantially larger gains than applying each component separately. Algorithm~\ref{alg:mce} summarizes the full MCE procedure.

\subsection{Network Architecture}
Implementation details for modality encoders, fusion modules, and decoders vary across PR applications (Sec.~\ref{exp:implement}). To improve generality and strengthen cross-modal recovery, we introduce a unified, attention-based reconstruction module that refines per-modality features prior to fusion.

We implement the reconstruction module with a Transformer that uses multi-head self-attention to model dependencies across modalities and reconstruct missing modality features. Let $\mathbf{h}_{multi} \in \mathcal{R}^{C \times M \times D}$ be the stacked multi-modal features regardless of modality presence; the reconstruction module is computed as
\begin{equation}
\begin{split}
    \mathbf{h}_{multi} & = \mathbf{h}_{multi} + E_{pos} \\
    \mathbf{h}_{multi} & = f_{MHSA}(\mathbf{h}_{multi},\mathbf{h}_{multi},\mathbf{h}_{multi}) \\
    \mathbf{h}_{multi} & = f_{FFN}(\mathbf{h}_{multi}) + \mathbf{h}_{multi} \\
\end{split}
\end{equation}
where $f_{MHSA}(\cdot)$ denotes multi-head self-attention~\cite{vaswani2017attention} and $f_{FFN}(\cdot)$ denotes a two-layer feed-forward network. $E_{pos} \in \mathcal{R}^{1 \times M \times D}$ is a learnable positional encoding that captures spatial structure and implicit modality relationships. For absent modalities, we input a zeroed feature at the corresponding position; using $E_{pos}$ as a learned mask token prevents the model from treating missing placeholders as observed evidence, and allows the Transformer to learn how to interpret absence during reconstruction.

The reconstruction module is applied before the fusion stage to complete and enrich per-modality features. Because it only augments feature representations and does not change the downstream fusion operator, the module is \textit{fusion-agnostic} and can be combined with a wide range of fusion strategies.


\section{Experiments}
\label{sec:experiment}

\subsection{Datasets and Pre-processing}
We evaluate the performance of MCE on four publicly available multi-modal benchmarks, each involving distinct tasks: urban scene segmentation (nuScenes), 3D brain tumor segmentation (BraTS2020), emotion recognition (IEMOCAP), and digit recognition (AudiovisionMNIST).

\textbf{nuScenes}~\cite{caesar2020nuscenes} is an urban-scene dataset made available for non-commercial use. The sensor suite includes 6 \textit{\textbf{C}ameras}, 5 \textit{\textbf{R}ADAR} units, and 1 \textit{\textbf{L}iDAR}. The cameras are arranged to capture different perspectives: front, front-left, front-right, back-left, back, and back-right. While the RADAR units cover front, left, right, back-left, and back-right positions. We use the official training and validation split in~\cite{philion2020lift}, consisting of 28,130 and 6,019 samples, respectively. 15 non-position items and 5 position items are all used as the raw RADAR data. Both LiDAR and RADAR are pre-processed to produce bird's-eye-view (BEV) feature maps, as described in~\cite{harley2023simple}. The segmentation task involves identifying vehicle pixels, aggregating all vehicle subclasses (\textit{e.g.}, bicycle, bus, car) into a single superclass.

\textbf{BraTS2020}~\cite{menze2014multimodal,bakas2017advancing,bakas2018identifying} (brain tumor segmentation challenge) consists of MRI scans in four modalities: \textit{\textbf{T1}}, \textit{\textbf{T1c}}, \textit{\textbf{Flair}}, and \textit{\textbf{T2}}. We use 369 labeled cases split into 219 for training, 50 for validation, and 100 for testing, as per the challenge convention. Tumor labels are grouped into whole tumor (WT), tumor core (TC), and enhancing tumor (ET). The volumes are resampled to a $1mm^3$ resolution, and random crops of $80 \times 80 \times 80$ voxels are used during training.

\textbf{IEMOCAP}~\cite{busso2008iemocap} is a multi-modal emotion recognition dataset that includes \textit{\textbf{A}coustic}, \textit{\textbf{V}isual} and \textit{\textbf{L}exical} modalities collected from five dyadic conversation sessions. We use the data split of 4,446/3,899/3,696 utterances for training, validation, and testing, respectively, focusing on four emotion categories: happiness, anger, sadness, and neutral. Acoustic features are extracted using the OpenSMILE toolkit (IS13 ComParE configuration)~\cite{eyben2010opensmile}; visual features are obtained from a DenseNet~\cite{huang2017densely} pre-trained on the Facial Expression Recognition Plus (FER+) corpus; lexical features are derived from a pretrained BERT-large model~\cite{devlin2019bert}.

\textbf{AudiovisionMNIST}~\cite{vielzeuf2018centralnet} contains paired \textit{\textbf{I}mage} and \textit{\textbf{A}udio} modalities. The image inputs consist of MNIST digits~\cite{lecun1998gradient} (size $28 \times 28$), while audio samples are sourced from the Free Spoken Digits Dataset~\footnote{\url{https://github.com/Jakobovski/free-spoken-digit-dataset}}. The dataset contains 1,500 paired samples, with 70\% assigned for training and the remainder for validation.

\begin{table}[t]
\footnotesize
\centering
\caption{Implementation details on evaluation datasets.}
\label{tab:implementation}
\setlength{\tabcolsep}{0.5mm}
\begin{tabular}{ccccc}
    \hline
    Item & nuScenes~\cite{caesar2020nuscenes} & BraTS2020~\cite{menze2014multimodal,bakas2017advancing,bakas2018identifying} & IEMOCAP~\cite{busso2008iemocap} & AudiovisionMNIST~\cite{vielzeuf2018centralnet} \\
    \hline
    Epoch/Iteration & 50k iters & 400 epochs & 60 epochs & 2.5k iters \\
    Batch size & 1 & 1 & 128 & 128 \\
    Optimizer & AdamW & AdamW & Adam & Adam\\
    Scheduler & OneCycle~\cite{smith2019super} & Poly decay ($p=0.9$) & Step decay & Step decay \\
    Learning rate & $3 \times 10^{-4}$ & $2 \times 10^{-4}$ & $2 \times 10^{-4}$ & $1 \times 10^{-3}$ \\
    Evaluation & IoU & Dice & Accuracy & Accuracy \\
    Device & RTX4090 & RTX4090 & RTX4090 & GTX1080Ti \\
    \hline
\end{tabular}
\end{table}

\begin{table}[t]
\footnotesize
\centering
\caption{Component ablation study. The missing rates during training are (0.2, 0.5, 0.8) for nuScenes (C, L, R) and IEMOCAP (A, L, V). The seed for training on nuScenes is set to 125. Results show average performance across all 7 multi-modal combinations.}
\label{tab:ablation}
\setlength{\tabcolsep}{0.6mm}
\begin{tabular}{cccccc|ccc}
    \hline
    No. & $\mathcal{A}$ & $\mathcal{B}$ & $\mathcal{L}_{single}$ & $\mathcal{L}_{sub}$ & $\mathcal{L}_{aux}$ & nuScenes (IoU) & IEMOCAP (Accuracy) & IEMOCAP (95\% CI)\\
    \hline
    a. & & & & & & 31.24 & $53.90{\pm 0.86}$ & [53.24, 54.55] \\
    \hdashline[3pt/3pt]
    b. & & & \ding{52} & & & 32.64 & $55.09{\pm 1.88}$ & [53.65, 56.54] \\
    c. & & & & \ding{52} & & 34.17 & $57.99{\pm 2.53}$ & [56.04, 59.94] \\
    d. & & & & & \ding{52} & 31.92 & $54.89{\pm 2.20}$ & [53.20, 56.58] \\
    e. & & & \ding{52} & \ding{52} & & 35.85 & $58.70{\pm 1.64}$ & [57.53, 59.87] \\
    f. & & & \ding{52} & & \ding{52} & 33.34 & $55.71{\pm 2.00}$ & [54.28, 57.14] \\
    g. & & & & \ding{52} & \ding{52} & 35.47  & $58.37{\pm 1.28}$ & [57.46, 59.28] \\
    h. & & & \ding{52} & \ding{52} & \ding{52} & 36.56 & $58.85{\pm 1.14}$ & [58.03, 59.66] \\
    \hdashline[3pt/3pt]
    i. & \ding{52} & & \ding{52} & \ding{52} & \ding{52} & 37.47 & $59.26{\pm 1.48}$ &  [58.20, 60.32] \\
    j. & & \ding{52} & \ding{52} & \ding{52} & \ding{52} & 38.15 & $60.48{\pm 1.25}$ & [59.59, 61.37] \\
    \hdashline[3pt/3pt]
    k. & \ding{52} & \ding{52} & \ding{52} & & & 36.28 & $56.93{\pm 1.28}$ & [56.01, 57.85] \\
    l. & \ding{52} & \ding{52} & & & \ding{52} & 34.88 & $55.46{\pm 1.75}$ & [54.21, 56.71] \\
    \hdashline[3pt/3pt]
    m. & \ding{52} & \ding{52} & \ding{52} & \ding{52} & \ding{52} & 39.80 & $60.93{\pm 1.39}$ & [59.94, 61.92] \\
    \hline
\end{tabular}
\end{table}

\subsection{Implementation Details}
\label{exp:implement}

For nuScenes, we assume data synchronization across sensors, with known intrinsics and relative poses. The 3D metric span is defined as $100m \times 10m \times 100m$ (left/right, up/down, and forward/backward), with voxel lengths set to $0.5m \times 1.25m \times 0.5m$, discretized at a resolution of $200 \times 8 \times 200$. The EfficientNet~\cite{tan2019efficientnet} backbone is used to encode RGB images into 2D feature maps, which are then lifted to 3D coordinates and collapsed along the vertical dimension to yield high-dimensional BEV feature maps~\cite{harley2023simple}. Simple concatenation is employed to fuse the modality feature maps, and two convolutional layers are used as the segmentation task head.

For BraTS2020, we use 3D U-Net to extract features from each modality and predict segmentation results. The mmFormer~\cite{zhang2022mmformer} fusion backbone is used for all models.

For IEMOCAP, independent LSTM networks~\cite{sak2014long} with max-pooling are used to generate acoustic and visual embeddings. A TextCNN~\cite{chen2015convolutional} is used to get lexical embedding. These embeddings are concatenated for multi-modal fusion. The decoder consists of two fully connected (FC) layers with ReLU activation and batch normalization, followed by another FC layer.

For AudiovisionMNIST, we use LeNet5~\cite{lecun1998gradient} as the encoder for the image modality and MFCCs~\cite{tzanetakis2002musical} as the encoder for the audio modality. The two modality features are fused using concatenation, and a fully connected layer follows. The decoder is built with a FC layer with ReLU activation, followed by another FC layer.

During evaluation, we report average performance across all possible multi-modal input combinations. For example, given the modality set (C, L, R), the results are aggregated over all seven test conditions: (C), (L), (R), (C, L), (C, R), (L, R), and (C, L, R). More implementation details are provided in Tab.~\ref{tab:implementation}.

\subsection{Component Analysis}
\label{sec:ablation}

\begin{figure}[t]
    \centering
    \includegraphics[width=\linewidth]{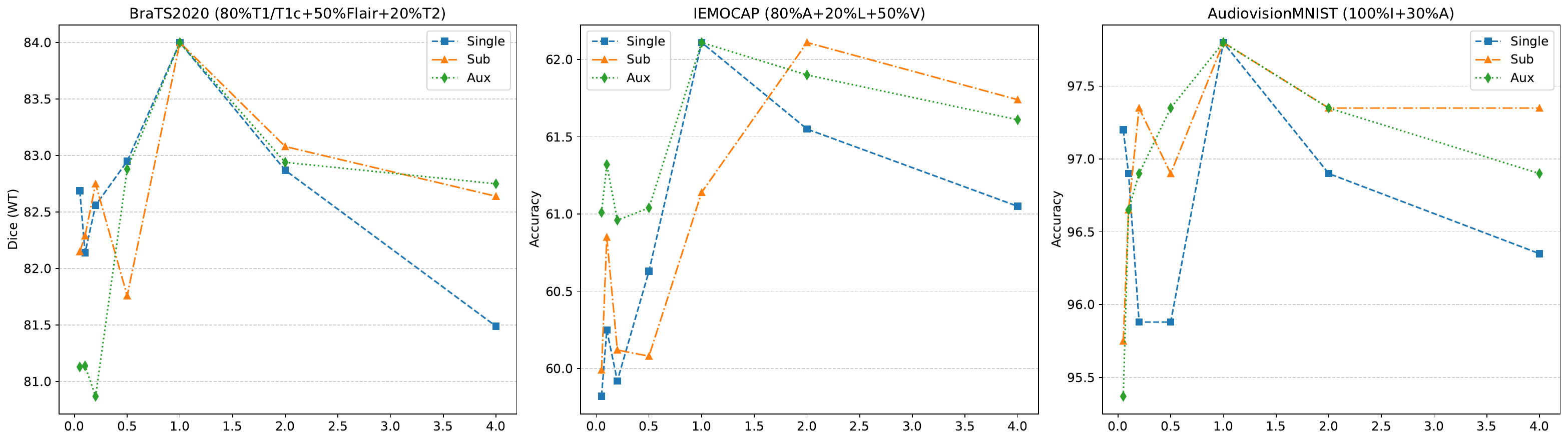}
    \vspace{-6mm}
    \caption{Performance across different hyperparameter configurations on three datasets.}
    \label{fig:hyper_independent}
\end{figure}

\begin{figure}[t]
    \centering
    \includegraphics[width=\linewidth]{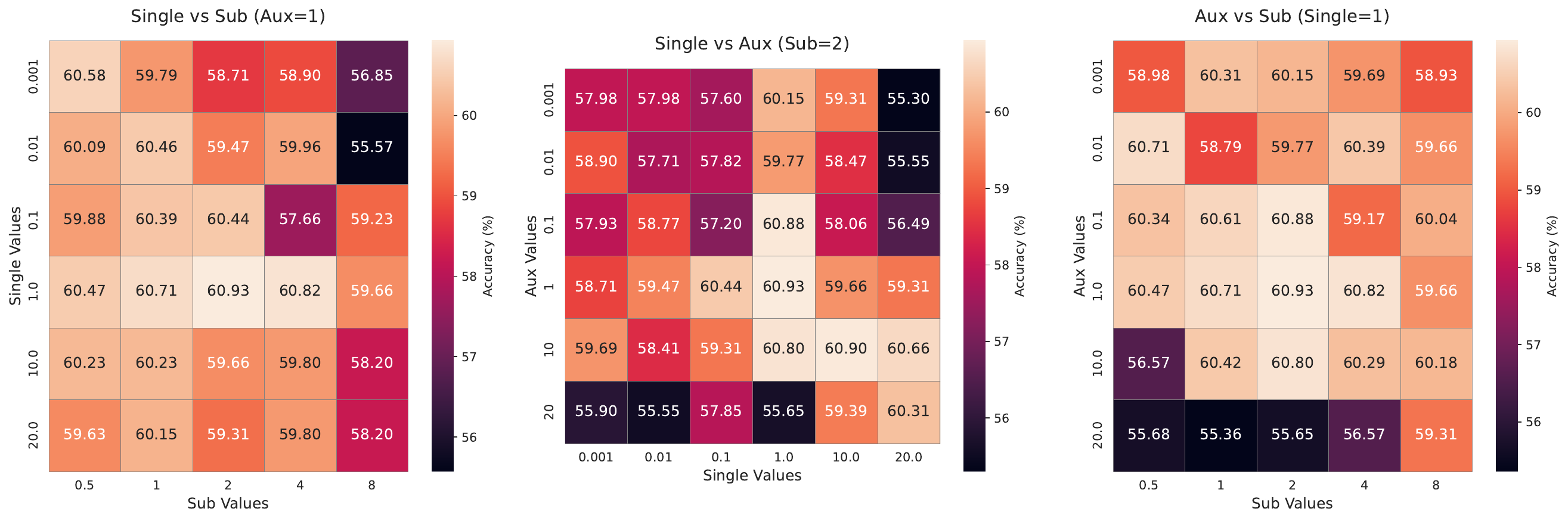}
    \vspace{-6mm}
    \caption{Hyperparameter interaction analysis on IEMOCAP.}
    \label{fig:hyper_heatmap}
\end{figure}

To thoroughly evaluate the proposed MCE framework, we conduct ablation studies and hyperparameter analyses to assess the contribution of each component and their interactions.

\textbf{Component contributions.} Tab.~\ref{tab:ablation} presents the ablation results, revealing several key insights. First, there is a clear, progressive improvement in performance from the baseline (a) to the full model (m), confirming the necessity of each proposed component. Second, comparing individual RCE components (b, c, d) with their combinations (e, f, g, h) reveals that the three supervisory signals within RCE are complementary. While using each signal individually leads to some improvement, their combination provides significantly better results ($e.g.$, a +2.39\% IoU increase on nuScenes and +0.86\% accuracy on IEMOCAP compared to the best dual combination (e)). This highlights the effectiveness of our multi-faceted representation enhancement strategy.

Most importantly, the results underscore the \textit{synergistic effect} between LCE and RCE. When both the global factor $\mathcal{A}$ (i) and the batch-level factor $\mathcal{B}$ (j) are incorporated on top of the complete RCE foundation (h), substantial performance improvements are observed. The fully integrated MCE framework (m), combining both LCE and RCE, achieves the highest performance (39.80 IoU, 60.93\% accuracy). In contrast, configurations (k) and (l), where LCE ($\mathcal{A} + \mathcal{B}$) is applied without the complete set of RCE supervisions, result in suboptimal performance, significantly underperforming the full model (m) ($e.g.$, -3.52 IoU and -5.47 accuracy for (k)). These findings clearly show that LCE alone does not suffice; it functions as a dynamic facilitator that enhances the effects of RCE, leading to a more balanced and effective optimization process.

\begin{figure}[t]
    \centering
    \includegraphics[width=0.9\linewidth]{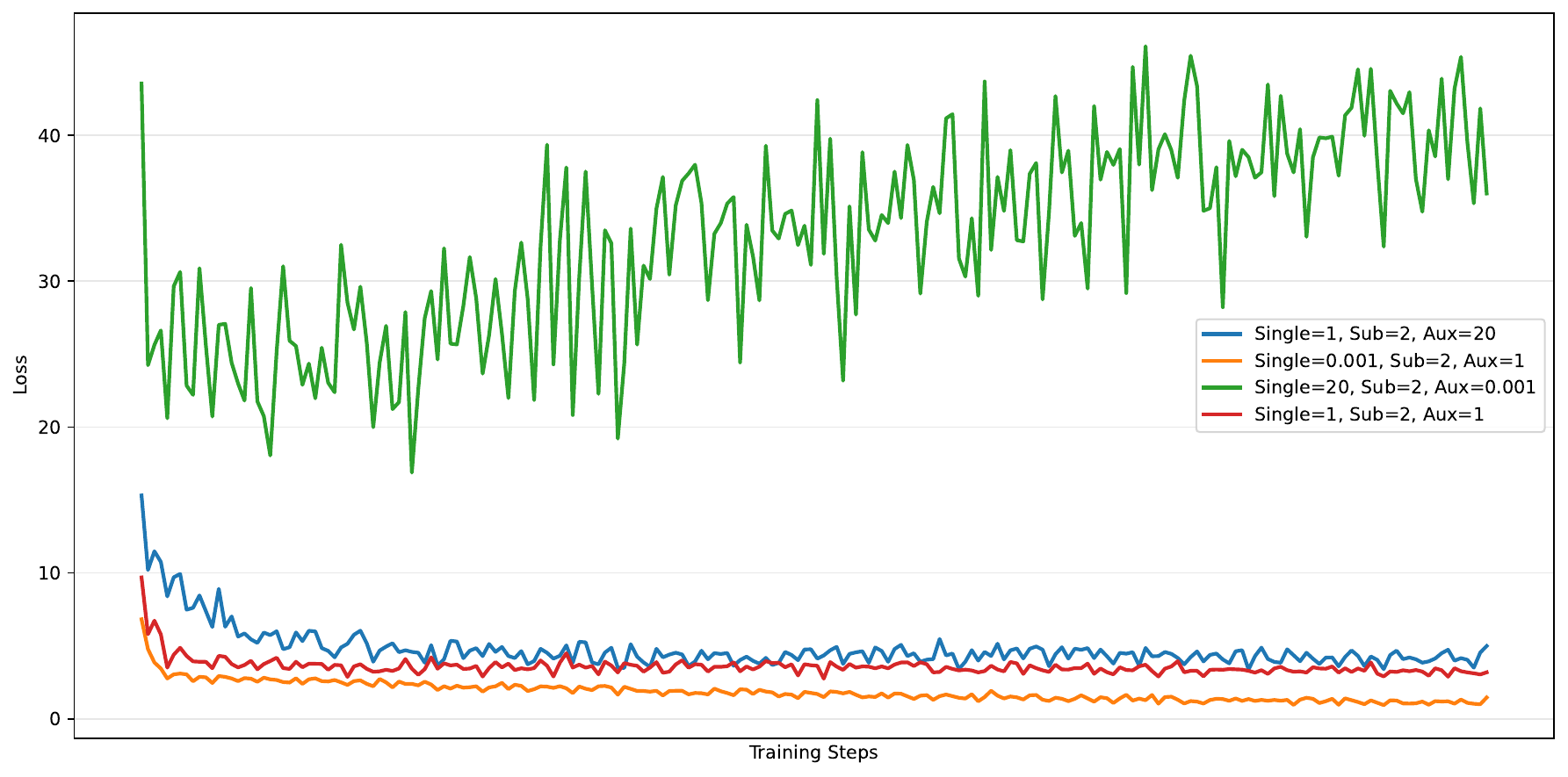}
    \vspace{-2mm}
    \caption{Training dynamics under different hyperparameter configurations on IEMOCAP.}
    \label{fig:hyper_loss}
\end{figure}

\textbf{Hyperparameter analysis.} The robustness of the MCE framework is further demonstrated in Fig.~\ref{fig:hyper_independent}, where performance remains stable across a wide range of hyperparameter values ($\lambda_{single}$, $\lambda_{sub}$, $\lambda_{aux}$) on three different datasets.

To further investigate the interactions between these hyperparameters and the coupling between the components they weight, we present interaction heatmaps for the IEMOCAP dataset in Fig.~\ref{fig:hyper_heatmap}. The results show a sizeable, warm-colored region of high performance, suggesting that our method is robust to the specific choice of hyperparameters within a reasonable range. More importantly, the optimal performance depends on a specific combination of parameters, which highlights the complex interactions between different loss terms.

The training dynamics in Fig.~\ref{fig:hyper_loss} provide additional insights, which presents the training loss curves under different hyperparameter settings. The optimal setting (1, 2, 1) shows a smooth, stable convergence trajectory, while other settings, such as (1, 2, 20) (over-weighted $\mathcal{L}_{aux}$), result in more severe oscillations than optimal setting, indicating optimization instability due to conflicting gradients. Conversely, settings like (0.001, 2, 1) (under-weighted $\mathcal{L}_{single}$) show an initial rapid decrease in loss but fail to fully converge by the end of training, suggesting an insufficient learning signal. These dynamics visually demonstrate the optimization implications of hyperparameter interactions: only a balanced combination ensures stable and efficient training, confirming the synergistic interplay between LCE and RCE.


\begin{table*}[t]
\footnotesize
\centering
\caption{Quantitative comparison (IoU) against different methods on nuScenes when training with different missing rates of (C, L, R). The fusion strategy, for method does not point out the unified fusion strategy in the original paper, is simple concatenation followed by convolutional layers. The evaluation results are the average performance across all 7 multi-modal combinations. The seed is set as 125.}
\label{tab:nuscenes}
\begin{tabular}{ccccccc}
    \hline
    Method & a & b & c & d & e & f \\
    \hline
    PMR~\cite{fan2023pmr} & 15.84 & 13.79 & 16.30 & 13.62 & 14.78 & 11.32 \\
    MBT-Sample~\cite{wei2024enhancing} & 8.37 & 10.93 & 9.65 & 7.69 & 11.12 & 10.18 \\
    \hdashline[3pt/3pt]
    ShaSpec~\cite{wang2023multi} & 31.95 & 30.80 & 32.25 & 29.88 & 30.39 & 27.36 \\
    RedCore~\cite{sun2024redcore} & 34.12 & 32.45 & 33.97 & 31.04 & 32.77 & 30.54 \\
    \hdashline[3pt/3pt]
    MCE &  \textbf{39.80} & \textbf{37.17} & \textbf{40.46} & \textbf{39.39} & \textbf{37.73} & \textbf{37.50} \\    
    \hline
\end{tabular}
\end{table*}

\begin{table*}[t]
\footnotesize
\centering
\caption{Quantitative comparison (WT/TC/ET) against different methods on BraTS2020 when training with different missing rates of (T1/T1c, Flair, T2). mmFormer~\cite{zhang2022mmformer} is selected as the baseline and the base backbone. The results represent the average performance across all 15 multi-modal combinations, based on 10 independent experimental runs.}
\label{tab:brats2020}
\setlength{\tabcolsep}{0.5mm}
\begin{tabular}{ccccccc}
    \hline
    WT & a & b & c & d & e & f \\
    \hdashline[3pt/3pt]
    baseline & $81.72_{\pm 0.87}$ & $81.68_{\pm 0.62}$ & $81.61_{\pm 0.95}$ & $81.25_{\pm 0.53}$ & $80.65_{\pm 0.68}$ & $80.68_{\pm 1.04}$ \\
    PMR~\cite{fan2023pmr} & $82.42_{\pm 0.87}$ & $82.38_{\pm 0.63}$ & $81.89_{\pm 0.52}$ & $82.05_{\pm 0.95}$ & $81.02_{\pm 0.41}$ & $80.97_{\pm 0.76}$ \\
    MBT-Sample~\cite{wei2024enhancing} & $83.10_{\pm 0.45}$ & $83.24_{\pm 0.62}$ & $82.69_{\pm 0.33}$ & $83.19_{\pm 0.57}$ & $82.65_{\pm 0.81}$ & $81.94_{\pm 0.39}$ \\
    PASSION~\cite{shi2024passion} & $83.37_{\pm 0.52}$ & $83.51_{\pm 0.63}$ & $82.93_{\pm 0.51}$ & $83.34_{\pm 0.45}$ & $82.65_{\pm 0.78}$ & $82.59_{\pm 0.27}$ \\
    MCE & $\mathbf{84.00_{\pm 0.15}}$ & $\mathbf{84.35_{\pm 0.57}}$ & $\mathbf{83.66_{\pm 0.35}}$ & $\mathbf{83.57_{\pm 0.05}}$ & $\mathbf{83.21_{\pm 0.06}}$ & $\mathbf{82.91_{\pm 0.17}}$ \\
    
    \hline
    TC & a & b & c & d & e & f \\
    \hdashline[3pt/3pt]
    baseline & $69.41_{\pm 0.76}$ & $67.89_{\pm 0.49}$ & $70.05_{\pm 0.91}$ & $68.63_{\pm 0.57}$ & $68.16_{\pm 0.82}$ & $68.22_{\pm 0.45}$ \\
    PMR~\cite{fan2023pmr} & $70.28_{\pm 0.68}$ & $68.27_{\pm 0.92}$ & $70.41_{\pm 0.57}$ & $69.43_{\pm 0.84}$ & $68.35_{\pm 0.49}$ & $68.03_{\pm 1.02}$ \\
    MBT-Sample~\cite{wei2024enhancing} & $70.62_{\pm 0.48}$ & $71.35_{\pm 0.71}$ & $70.17_{\pm 0.56}$ & $70.27_{\pm 0.63}$ & $69.79_{\pm 0.42}$ & $68.80_{\pm 0.68}$ \\
    PASSION~\cite{shi2024passion} & $71.67_{\pm 0.36}$ & $71.49_{\pm 0.59}$ & $70.85_{\pm 0.42}$ & $70.40_{\pm 0.68}$ & $69.92_{\pm 0.23}$ & $69.36_{\pm 0.74}$ \\
    MCE & $\mathbf{72.62_{\pm 0.33}}$ & $\mathbf{72.34_{\pm 0.15}}$ & $\mathbf{71.82_{\pm 0.18}}$ & $\mathbf{71.93_{\pm 0.39}}$ & $\mathbf{71.04_{\pm 0.58}}$ & $\mathbf{70.67_{\pm 0.48}}$ \\

    \hline
    ET & a & b & c & d & e & f \\
    \hdashline[3pt/3pt]
    baseline & $51.65_{\pm 0.98}$ & $50.21_{\pm 0.64}$ & $52.23_{\pm 0.71}$ & $51.87_{\pm 0.59}$ & $51.07_{\pm 0.86}$ & $51.92_{\pm 1.07}$ \\
    PMR~\cite{fan2023pmr} & $52.67_{\pm 0.73}$ & $50.75_{\pm 0.45}$ & $51.72_{\pm 0.61}$ & $52.88_{\pm 0.98}$ & $51.72_{\pm 0.54}$ & $50.78_{\pm 0.89}$ \\
    MBT-Sample~\cite{wei2024enhancing} & $52.30_{\pm 0.53}$ & $52.74_{\pm 0.37}$ & $52.17_{\pm 0.49}$ & $53.26_{\pm 0.66}$ & $52.97_{\pm 0.58}$ & $51.94_{\pm 0.72}$ \\
    PASSION~\cite{shi2024passion} & $52.93_{\pm 0.61}$ & $53.21_{\pm 0.33}$ & $52.73_{\pm 0.47}$ & $53.15_{\pm 0.55}$ & $52.27_{\pm 0.39}$ & $52.08_{\pm 0.29}$ \\
    MCE & $\mathbf{54.52_{\pm 0.53}}$ & $\mathbf{54.38_{\pm 0.22}}$ & $\mathbf{53.80_{\pm 0.52}}$ & $\mathbf{54.21_{\pm 0.59}}$ & $\mathbf{53.16_{\pm 0.44}}$ & $\mathbf{52.70_{\pm 0.33}}$ \\

    \hline
\end{tabular}
\end{table*}

\begin{table*}[t]
\footnotesize
\centering
\caption{Quantitative comparison (Accuracy) against different methods on IEMOCAP when training with missing rate settings of (A, L, V). MMIN~\cite{zhao2021missing} is selected as the baseline and the base backbone. The results represent the average performance across all 7 multi-modal combinations, based on 10 independent experimental runs.}
\setlength{\tabcolsep}{0.5mm}
\label{tab:iemocap}
\begin{tabular}{c|cccccccc}
    \hline
    Method & a & b & c & d & e & f \\
    \hline
    baseline & $49.72_{\pm 4.52}$ & $56.36_{\pm 9.46}$ & $43.36_{\pm 6.35}$ & $60.31_{\pm 3.02}$ & $38.98_{\pm 8.84}$ & $33.26_{\pm 3.14}$ \\
    UniMF~\cite{huan2023unimf} & $49.24_{\pm 2.03}$ & $57.60_{\pm 3.04}$ & $55.86_{\pm 3.17}$ & $58.45_{\pm 4.08}$ & $59.28_{\pm 5.76}$ & $54.67_{\pm 3.82}$ \\
    MBT-Sample~\cite{wei2024enhancing} & $57.13_{\pm 0.95}$ & $60.75_{\pm 0.88}$ & $59.99_{\pm 1.06}$ & $59.24_{\pm 1.69}$ & $61.24_{\pm 1.44}$ & $60.86_{\pm 0.94}$ \\
    RedCore~\cite{sun2024redcore} & $54.62_{\pm 1.32}$ & $59.25_{\pm 1.92}$ & $60.83_{\pm 0.76}$ & $45.89_{\pm 2.02}$ & $60.14_{\pm 1.71}$ & $45.83_{\pm 1.16}$ \\
    MCE & $\mathbf{60.93_{\pm 1.39}}$ & $\mathbf{62.11_{\pm 1.63}}$ & $\mathbf{61.55_{\pm 1.00}}$ & $\mathbf{63.75_{\pm 2.04}}$ & $\mathbf{63.36_{\pm 1.24}}$ & $\mathbf{62.59_{\pm 0.97}}$ \\
    \hline
\end{tabular}
\end{table*}

\begin{figure}[t]
    \centering
    \includegraphics[width=\linewidth]{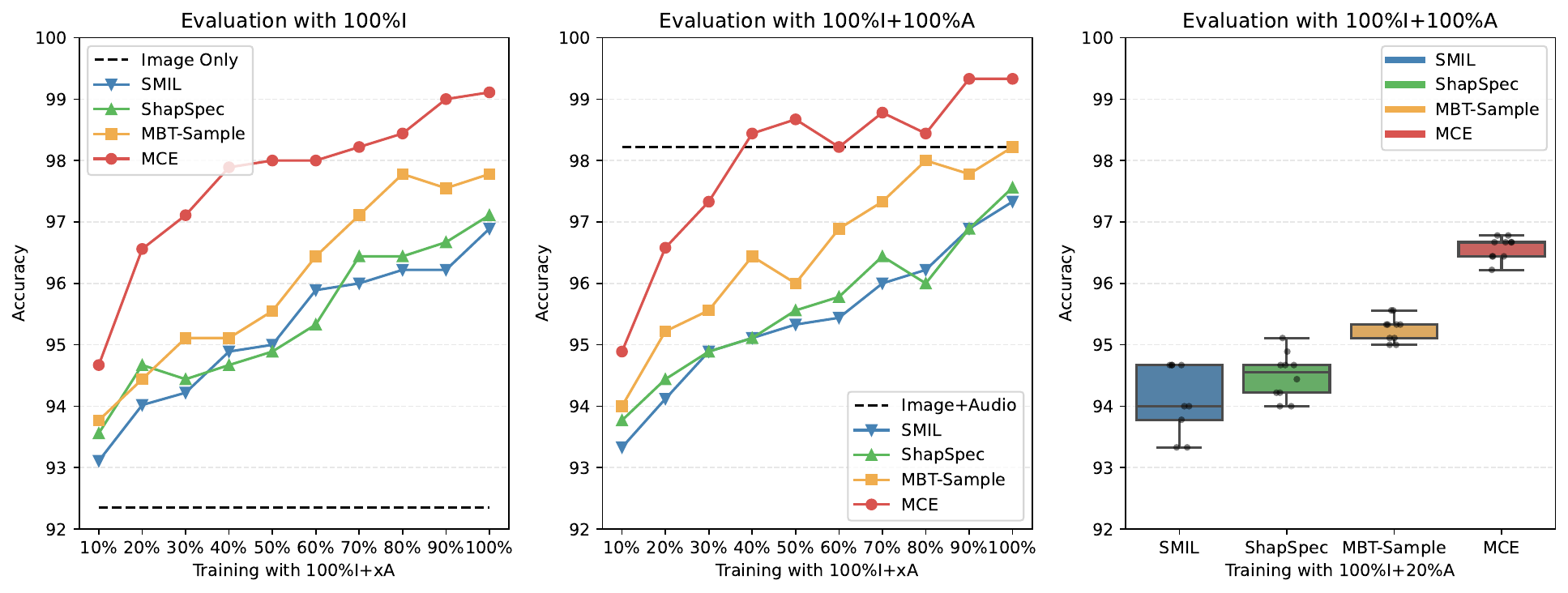}
    \vspace{-5mm}
    \caption{Quantitative comparison (Accuracy) on AudiovisonMNIST. All methods are trained with all images and different rate of audio, and are evaluated with \textit{Image Only} (left) or \textit{Image+Audio} (middle\&right). The results in box plot (right) are based on 10 independent experimental runs. \textit{Image Only} is a LeNet5~\cite{lecun1998gradient} trained with image only. \textit{Image+Audio} is a multi-modal model trained with both modalities.}
    \label{fig:audiovision}
\end{figure}

\subsection{Flexibility and Generalization Analysis}
\label{sec:main_results}

To evaluate the robustness and flexibility of MCE, we train it under highly imbalanced missing rates, selecting settings of 0.2 (mild), 0.5 (moderate), and 0.8 (severe). This results in six representative combinations of imbalanced missing rates: a. (0.2, 0.5, 0.8); b. (0.2, 0.8, 0.5); c. (0.5, 0.2, 0.8); d. (0.5, 0.8, 0.2); e. (0.8, 0.2, 0.5); f. (0.8, 0.5, 0.2). We compare with methods addressing imbalanced learning (PMR~\cite{fan2023pmr}, MBT-Sample~\cite{wei2024enhancing}), incomplete learning (ShaSpec~\cite{wang2023multi}, UniMF~\cite{huan2023unimf}, SMIL~\cite{ma2021smil}), and both aspects (RedCore~\cite{sun2024redcore}, PASSION~\cite{shi2024passion}). Results across different datasets are summarized in Tab.~\ref{tab:nuscenes},~\ref{tab:brats2020},~\ref{tab:iemocap}, and Fig.~\ref{fig:audiovision}.

The results confirm that MCE consistently outperforms other methods. On the nuScenes dataset, both PMR and MBT-Sample struggle with generalization. These methods predominantly rely on a basic fusion strategy—concatenation followed by convolutional layers—and focus mainly on addressing modality imbalance without directly handling missing modalities. While PMR slightly outperforms MBT-Sample by applying additional supervision to each modality, both methods fail to fully address the issue of missing modalities. ShaSpec, which focuses primarily on available modalities, also does not directly mitigate modality imbalance.

RedCore, which targets both modality imbalance and missingness, uses a feature imputation strategy to reconstruct missing modalities, improving performance over the previous methods. However, it still relies mainly on mean squared error (MSE) between imputed and original modality features to regularize the learning process under imbalance. In contrast, MCE introduces several key advantages: i) It uses more authoritative supervision signals, incorporating ground-truth and pre-trained single-modality models to serve as evaluation criteria and modality-specific performance upperbounds; ii) It enhances feature diversity and semantic representation through auxiliary tasks such as subset prediction and feature completion; iii) It leverages a powerful Transformer-based architecture to model inter-modal dependencies and reconstruct missing features. These innovations allow MCE to achieve significantly more robust and competitive performance.

On the other three datasets, baseline methods already incorporate mechanisms for completing missing modalities. These methods perform reasonably well on tasks for which they were originally designed. Nevertheless, MCE consistently maintains high performance, outperforming other approaches in nearly all comparisons. Moreover, across 10 independent experimental runs, MCE shows minimal performance fluctuation, indicating statistically robust and stable results.

\subsection{Case Study}
\label{sec:case_study}

\begin{table}[t]
\footnotesize
\centering
\caption{Batch-level Shapley value contribution analysis ($MR$ = missing rate, $BS$ = batch size).}
\setlength{\tabcolsep}{2mm}
\label{tab:batch_analysis}
\begin{tabular}{c|c|cccc|ccc}
    \hline
    Stage & Dataset & \multicolumn{4}{c|}{\makecell{BraTS2020\\$MR$=(0.8,0.5,0.2), $BS$=1}} & \multicolumn{3}{c}{\makecell{IEMOCAP\\$MR$=(0.2,0.5,0.8), $BS$=128}} \\
    \hline
    \multirow{5}{*}{\makecell{Early\\epoch}} & Modality & Flair & T1c & \st{T1} & T2 & A & L & V \\
    & Number & 1 & 1 & 0 & 1 & 101 & 65 & 20 \\
    & $\mathcal{U}_m$ & 0.8216 & 0.8051 & 0.8018 & 0.8283 & 91 & 26 & 12 \\
    & $\phi_m$ & 0.3053 & 0.2146 & 0.0000 & 0.2013 & 23.6667 & 11.6667 & 4.0000 \\
    & $\mathcal{B}_m$ & 0.5163 & 0.5905 & 0.0000 & 0.6271 & 0.6667 & 0.2205 & 0.4000 \\
    \hline
    \multirow{5}{*}{\makecell{Late\\epoch}} & Modality & \st{Flair} & T1c & T1 & \st{T2} & A & L & V \\
    & Number & 0 & 1 & 1 & 0 & 105 & 79 & 25 \\
    & $\mathcal{U}_m$ & 0.8216 & 0.8051 & 0.8018 & 0.8283 & 91 & 29 & 8 \\
    & $\phi_m$ & 0.0000 & 0.3514 & 0.3473 & 0.0000 & 59.6667 & 34.6667 & 10.6667 \\
    & $\mathcal{B}_m$ & 0.0000 & 0.4537 & 0.4545 & 0.0000 & 0.2984 & -0.0717 & -0.1067 \\
    \hline
\end{tabular}
\end{table}

\begin{figure}[t]
    \centering
    \includegraphics[width=\linewidth]{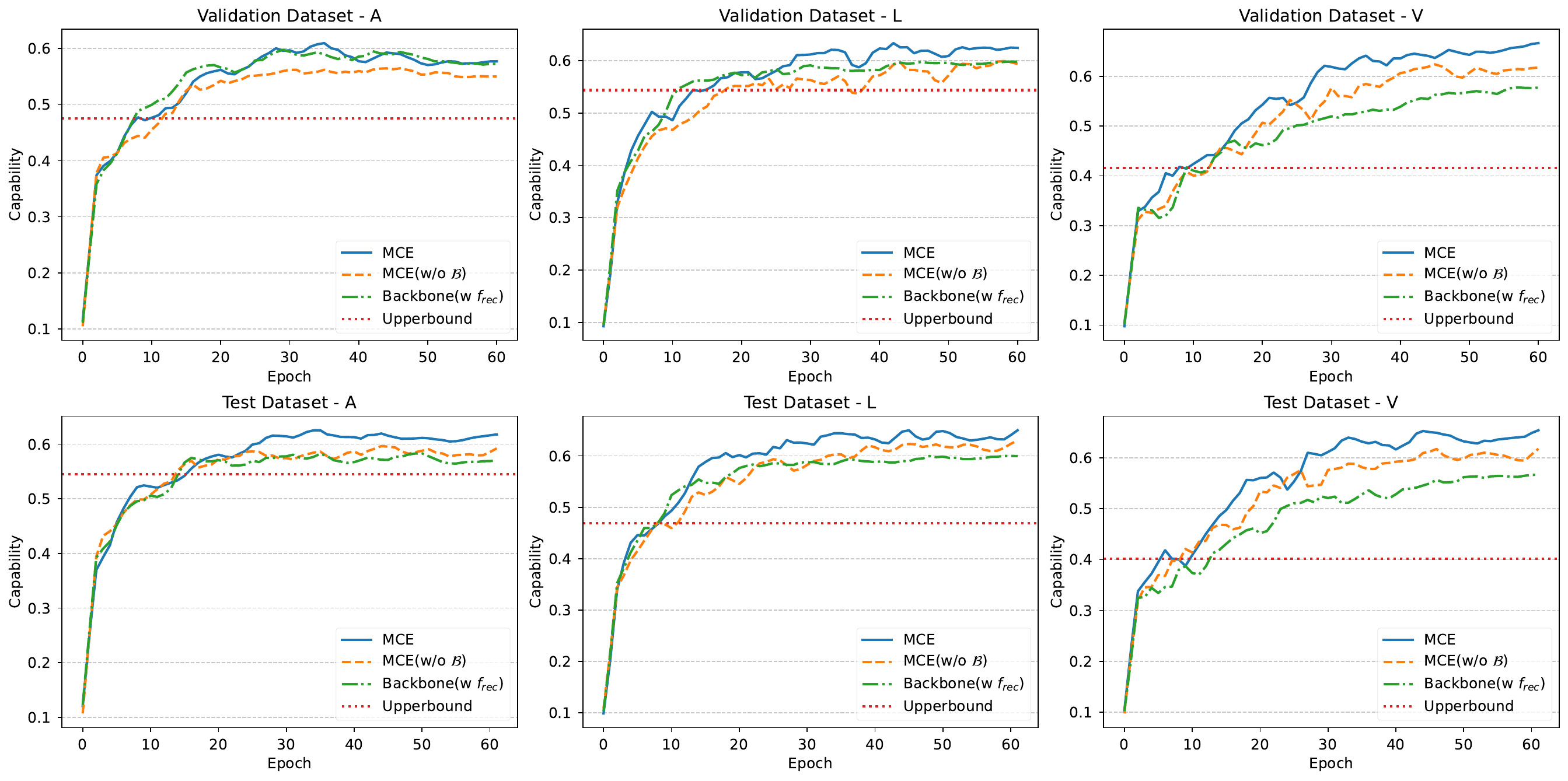}
    \caption{Evolution of Modality Capability on IEMOCAP. This figure tracks the capability (measured by the performance upper bound $\mathcal{U}_m$) of Acoustic (A), Lexical (L), and Visual (V) modalities throughout the training process on both validation and test sets. The trajectories compare our full MCE model (blue, with $\mathcal{B}$), an ablated version of our model without the $\mathcal{B}$ factor (orange), and the baseline backbone (green). The \textit{Upperbound} is the performance of a model trained with complete single-modal data.}
    \label{fig:capability_evolution}
\end{figure}

\textbf{Analysis of Shapley value contributions.} We begin by presenting a batch-level analysis to understand the dynamic behavior of the Shapley value ($\phi_m$) and the learning state factor ($\mathcal{B}_m$). Tab.~\ref{tab:batch_analysis} tracks these values for specific batches from the BraTS2020 and IEMOCAP datasets at both early and late stages of training. The results reveal the core mechanism of our approach. In the early stage, modalities with high missing rates (\textit{e.g.}, \textit{T2} in BraTS2020) are identified as underperforming, receiving a high encouragement factor ($\mathcal{B}_{T2}=0.6271$) due to their low Shapley value ($\phi_{T2}=0.2013$).

The most compelling insights arise in the later stages, showcasing the sophisticated and context-aware behaviors of our model. In BraTS2020, the T1c and T1 modalities exhibit very similar Shapley values ($\phi_{T1c}=0.3514$ vs. $\phi_{T1}=0.3473$) and encouragement factors ($\mathcal{B}_{T1c}=0.4537$ vs. $\mathcal{B}_{T1}=0.4545$). This suggests a state of healthy competition and achieves \textit{competitive equilibrium}, where the model perceives both modalities as equally valuable but still underperforming relative to their potential, thus incentivizing them equally to drive concurrent improvement. In IEMOCAP, the $\mathcal{B}_m$ values for both $L$ and $V$ modalities are negative ($-0.0717$ and $-0.1067$), while only $A$ receives a positive factor ($0.2984$). This behavior is not an anomaly but a reflection of the model's precision to focus \textit{context-aware incentive}. It indicates that, for the specific samples in this batch, $L$ and $V$ are performing exceptionally well, which surpasses the model's expectation. And therefore no additional encouragement is required. As a result, their incentive factors are masked to zero, and all the model's incentivization resources are directed toward modality $A$. This demonstrates the model’s context-sensitive ability to allocate learning resources efficiently.

Fig.~\ref{fig:capability_evolution} also provides a compelling visual analysis of the core mechanism behind MCE's effectiveness. The trajectories clearly demonstrate that while all three models (baseline (backbone w $f_{rec}$), MCE w/o $\mathcal{B}$, and Full MCE) facilitate learning across all modalities, our proposed framework induces a fundamentally different and more robust dynamic. Crucially, the ablation model (MCE w/o $\mathcal{B}$), already shows a marked positive improvement in the final capability of all modalities compared to the baseline. It underscores the value of our representation enhancement tasks. However, the introduction of the $\mathcal{B}$ factor in the full MCE model delivers a further significant boost, most notably for the under-performing modalities. This indicates that the Shapley value-based $\mathcal{B}$ factor successfully diagnoses capability gaps and dynamically allocates stronger learning incentives to the modalities that need it most. Therefore, it successfully accelerates, balances and elevates the learning process to achieve superior overall performance.

These findings highlight how $\mathcal{B}_m$ functions as a dynamic, self-regulating signal that manages learning incentives. It promotes fair competition across modalities and performs precise, sample-aware resource allocation by focusing on the modality that needs the most attention, ensuring robustness through the masking mechanism.

\begin{table}[t]
\footnotesize
\centering
\caption{Quantitative evaluation of representation quality on the IEMOCAP test set.}
\setlength{\tabcolsep}{3mm}
\label{tab:representation_quality}
\begin{tabular}{c|cccc}
    \hline
    Model & \makecell{Intra-class\\Distance ($\downarrow$)} & \makecell{Inter-class\\Distance ($\uparrow$)} & \makecell{Intra-/Inter-\\Ratio ($\downarrow$)} & \makecell{Avg. Intra-class\\Cosine Consistency ($\uparrow$)} \\
    \hline
    Baseline & 38.61 & 30.79 & 1.25 & 0.64 \\
    MCE & 11.96 & 14.15 & 0.85 & 0.73 \\
    \hline
\end{tabular}
\end{table}

\textbf{Analysis of representation quality.} To validate the effectiveness of the RCE component in improving learned representations, we evaluate the quality of the fused features $\mathbf{h}_n$ using two metrics: \textit{Intra-class to Inter-class Distance Ratio} and \textit{Average Intra-class Cosine Consistency}. The results, shown in Tab.~\ref{tab:representation_quality}, strongly support the superior representation capabilities of our MCE framework.

A significant transformation in the feature space structure is observed. First, the intra-class distance is reduced by 69.0\% (from 38.61 to 11.96), indicating that samples from the same class are now much more tightly clustered in the latent space. Although the inter-class distance decreases slightly due to global feature compaction, the intra-/inter-ratio improves by 32.0\% (from 1.25 to 0.85), demonstrating enhanced separation between different classes. This leads to more discriminative features and improved classification decision boundaries. Additionally, the average intra-class cosine consistency increases by 14.1\% (from 0.64 to 0.73), suggesting that the model is learning more stable and consistent representations for each class. This increased consistency means that the model maps samples from the same category to more similar regions in the feature space, regardless of missing modalities, enhancing robustness against incomplete inputs.

These improvements are directly attributed to the multi-faceted supervision in the RCE component. The $\mathcal{L}_{single}$ loss ensures each unimodal encoder produces discriminative features, directly reducing intra-class variance. The $\mathcal{L}_{sub}$ loss encourages the fusion module to generate effective features from any modality subset, improving both inter-class separation and intra-class consistency. Lastly, the $\mathcal{L}_{aux}$ task builds a coherent cross-modal latent space through reconstruction, further enhancing the organization and stability of the learned representations. Together, these components result in a feature space that is more discriminative, consistent, and robust, providing a strong foundation for the performance gains seen in our main results.

\begin{figure}[t]
    \centering
    \includegraphics[width=\linewidth]{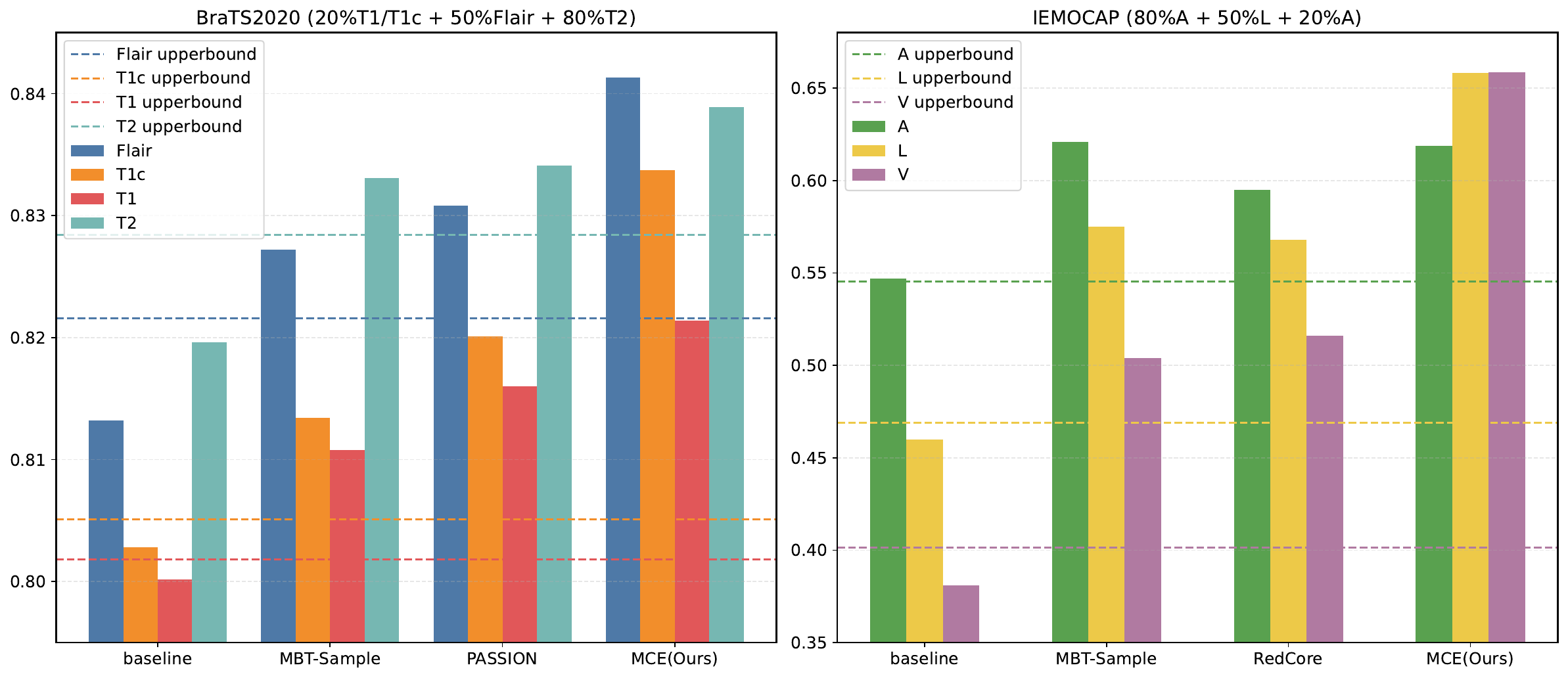}
    \caption{Modality capability comparison under (T1/T1c, Flair, T2)=(0.8, 0.5, 0.2) and (A, L, V)=(0.2, 0.5, 0.8) missing Rates on BraTS2020 and IEMOCAP, respectively. The \textit{Upperbound} is the performance of a model trained with complete single-modal data.}
    \label{fig:capability_comparison}
\end{figure}

\textbf{Modality capability analysis.} This section tests the core objective of our framework: addressing severely imbalanced missing rates. We evaluate the capability of encoders learned within different multi-modal frameworks by fixing them and training a new decoder using only that modality's data. The results, shown in Fig.~\ref{fig:capability_comparison}, reveal the exceptional capability of the MCE framework.

The most notable finding is the remarkable recovery of minority modalities. For IEMOCAP, the $V$ modality (with 80\% missing) is nearly ineffective in the baseline model, achieving a capability score of only 0.38. Other SOTA methods like MBT-Sample, which aims to enhance modal contributions, and RedCore, designed to balance inter-modal learning, show modest improvements. Our MCE framework achieves a dramatic 72.9\% relative increase, raising the $V$ modality's performance to 0.66. A similar recovery is seen for the \textit{T1/T1c} modalities (80\% missing) on BraTS2020. These results demonstrate that the Shapley-value-based incentive mechanism ($\mathcal{B}_m$) is particularly effective at rescuing modalities that would otherwise be neglected due to high absence rates. Moreover, MCE does not neglect the majority modalities. The $T2$ (BraTS2020) and $A$ (IEMOCAP) modalities, with low missing rates of 20\%, also achieve their highest capability under our framework. This indicates that MCE promotes a true equilibrium, enhancing all modalities rather than favoring only the minority ones. Finally, the results show that the multi-modal training process itself enables encoders to surpass their single-modal upperbounds. This suggests that inter-modal interactions, such as knowledge distillation and regularization during joint training, can create more powerful encoders than those trained in isolation. MCE effectively facilitates this exchange, enabling each modality to reach its theoretical potential.

These analyses conducted under challenging conditions provides conclusive evidence that MCE is designed to handle the core challenge of imbalanced missing rates. Its strength lies in its ability to prevent the collapse of minority modalities while ensuring balanced learning for all modalities, driving optimal performance across the board.


\begin{table}[t]
\footnotesize
\centering
\caption{Empirical analysis of computational and memory overhead on IEMOCAP dataset (Batch size=128; Missing rate (A, L, V)=(0.2, 0.5, 0.8)).}
\setlength{\tabcolsep}{1.3mm}
\label{tab:runtime}
\begin{tabular}{ccccc}
\hline
Method & Time (ms) & FLOPs (G) & GPU Memory (MB) & Accuracy \\
\hline
Baseline & 82.6 & 10.201 & 1425 & 52.54 \\
MCE (Exact SV) & 1046.9 & 10.480 & 1589 & 60.93 \\
MCE (MC-200 SV) & 232.1 & 10.480 & 1531 & 59.04 \\
\hline
\end{tabular}
\end{table}

\section{Discussion: Computational Considerations and Practical Trade-offs}
\label{sec:discussion}
The Shapley value calculation is justified by its unique benefits. Unlike heuristic approaches that rely solely on missing rates or importance measures, the Shapley value accounts for both individual modality contributions and their interactions within different coalitions. This comprehensive method enables more precise allocation of learning resources, particularly for underrepresented modalities that may otherwise be overlooked. However, despite its advantages, the Shapley value introduces both computational and memory overhead, as shown in Tab.~\ref{tab:runtime}. While the memory overhead increases from 1425 MB to 1589 MB, it is modest compared to the substantial increase in processing time. The exact Shapley value computation increases processing time by approximately $12 \times$ compared to baseline implementations. This is due to the exponential complexity ($O(2^M)$) of evaluating all possible modality subsets, which becomes prohibitive as the number of modalities increases.

We consider to employ the Monte Carlo sampling strategy, reducing the complexity to $O(K \cdot M)$ by considering $K$ random permutations of modality joining orders for each sample. In our experiments, $K=100$. The reasons are: i) providing a consistent and stable valuation across diverse modality subsets arising from different missing patterns; ii) reducing estimation variance by oversampling, ensuring that Shapley values yield smooth and reliable gradient signals; iii) optimizing efficiency by avoiding recomputation for identical sampling outcomes, instead using a weighted summation of the results. This approximation significantly reduces computational demands while maintaining competitive performance (59.04 vs. 60.93). For practical deployment, the decision should be based on the specific performance requirements of the application. 


\section{Conclusion}
\label{sec:conclusion}
This paper presents Modality Capability Enhancement (MCE), a principled framework that addresses imbalanced missing rates in incomplete multi-modal learning. The core innovation lies in the tightly coupled diagnosis–treatment loop: LCE dynamically identifies and incentivizes under-performing modalities through a dual-factor mechanism, while RCE translates these incentives into robust, cross-modally consistent features through subset prediction and auxiliary completion tasks.

The framework's principal strengths are threefold. First, it establishes a \textit{theoretically grounded} approach to imbalance mitigation through its game-theoretic formulation, providing rigorous quantification of modality contributions. Second, it maintains \textit{practical applicability} through its dual-factor design. The dataset-level factor $\mathcal{A}$ handles global missing rate imbalances while the batch-level factor $\mathcal{B}$ captures fine-grained learning states. Both of them operates independently of the specific fusion strategy. Third, it demonstrates \textit{holistic effectiveness} by simultaneously optimizing learning dynamics and representation quality, which enables significant recovery of underrepresented modalities while maintaining performance across diverse missing scenarios.

Despite these strengths, \textbf{MCE faces certain limitations}. The most significant of these is the computational overhead associated with Shapley value estimation, which could limit practical deployment in resource-constrained environments. Future work will focus on developing more efficient approximation methods, such as learned predictors of modality contributions or gradient-based estimation techniques. Additionally, the hyperparameters ($\lambda_{single}$, $\lambda_{sub}$, $\lambda_{aux}$) used in the experiments are based on simple tuning. Developing adaptive mechanisms to dynamically adjust these parameters according to real-time training progress would enhance both the framework’s practicality and ease of use. Another promising direction is to capture non-linear and synergistic interactions between modalities, moving beyond the current additive utility assumption. These advancements would transform MCE from an empirically successful solution into a more comprehensive, theoretically grounded framework.

In summary, MCE provides a practical and effective solution for handling real-world multi-modal data imperfections. We hope the framework's principles of synergistic learning and representation balancing will facilitate the development of more adaptable and efficient multi-modal learning systems, ultimately advancing robust pattern recognition in challenging real-world conditions.







\bibliographystyle{unsrt}
\bibliography{ref}

\end{document}